\documentclass[sigconf,screen=true,natbib=true,anonymous=false]{acmart}

\usepackage{multirow}
\usepackage{makecell}
\usepackage{cleveref}
\usepackage{graphicx}
\usepackage{arydshln}
\usepackage{enumitem}
\usepackage{fontawesome}
\usepackage{fancyvrb}

\DefineVerbatimEnvironment{CodeWithIcons}{Verbatim}{fontsize=\small,commandchars=\\\{\}}

\AtBeginDocument{%
  }

\copyrightyear{2026}
\acmYear{2026}
\setcopyright{cc}
\setcctype{by}
\acmConference[CIKM '26]{Proceedings of the 35th ACM International Conference on Information and Knowledge Management}{November 07--11, 2026}{Rome, Italy}
\acmBooktitle{Proceedings of the 35th ACM International Conference on Information and Knowledge Management (CIKM '26), November 07--11, 2026, Rome, Italy}
\acmDOI{10.1145/3799682.3840172}
\acmISBN{979-8-4007-2539-5/2026/11}
\acmSubmissionID{1006}

\begin{document}

\title{\textsc{Eval4Sim}: An Evaluation Framework for Persona Simulation}

\author{Eliseo Bao}
\email{eliseo.bao@udc.es}
\orcid{0009-0000-8457-1115}
\affiliation{
  \institution{IRLab, CITIC, Universidade da Coruña}
  \city{A Coruña}
  \country{Spain}
}

\author{Anxo Perez}
\email{anxo.pvila@udc.es}
\orcid{0000-0002-0480-006X}
\affiliation{
  \institution{IRLab, CITIC, Universidade da Coruña}
  \city{A Coruña}
  \country{Spain}
}

\author{Javier Parapar}
\email{javier.parapar@udc.es}
\orcid{0000-0002-5997-8252}
\affiliation{
  \institution{IRLab, CITIC, Universidade da Coruña}
  \city{A Coruña}
  \country{Spain}
}

\author{Xi Wang}
\email{xi.wang@sheffield.ac.uk}
\orcid{0000-0001-5936-9919}
\affiliation{
  \institution{University of Sheffield}
  \city{Sheffield}
  \country{United Kingdom}
}

\renewcommand{\shortauthors}{Eliseo Bao, Anxo Perez, Javier Parapar, and Xi Wang}

\begin{abstract}
Large Language Model personas, explicit profiles specifying a user's attributes, preferences, and behavioural tendencies, are increasingly used to simulate human conversations for user modelling, social reasoning, and behavioural analysis. Evaluating whether such simulations faithfully reflect human conversational behaviour is critical, yet current practice often relies on LLM-as-a-judge approaches that provide limited grounding in observable behaviour and produce opaque scalar scores. We present \textsc{Eval4Sim}, an evaluation framework that measures alignment between simulated and human conversations across three dimensions: \textit{adherence}, whether persona traits are recoverable from dialogue via dense retrieval; \textit{consistency}, whether a persona maintains a distinguishable stylistic identity via authorship verification; and \textit{naturalness}, whether conversations exhibit human-like turn-to-turn flow via dialogue NLI. Unlike optimization-oriented metrics, each dimension takes a human corpus as a reference baseline and penalizes deviations in both directions, distinguishing insufficient persona encoding from over-optimized, unnatural behaviour. The framework is corpus-agnostic: any persona-annotated conversational dataset can serve as the reference. Evaluated over ten simulation corpora, \textsc{Eval4Sim} surfaces systematic trade-offs invisible to single-score methods.
\end{abstract}




\acmDOI{}
\maketitle

\section{Introduction and Motivation}
\label{sec:introduction}

\begin{figure*}[t]
    \centering
    \includegraphics[width=\linewidth]{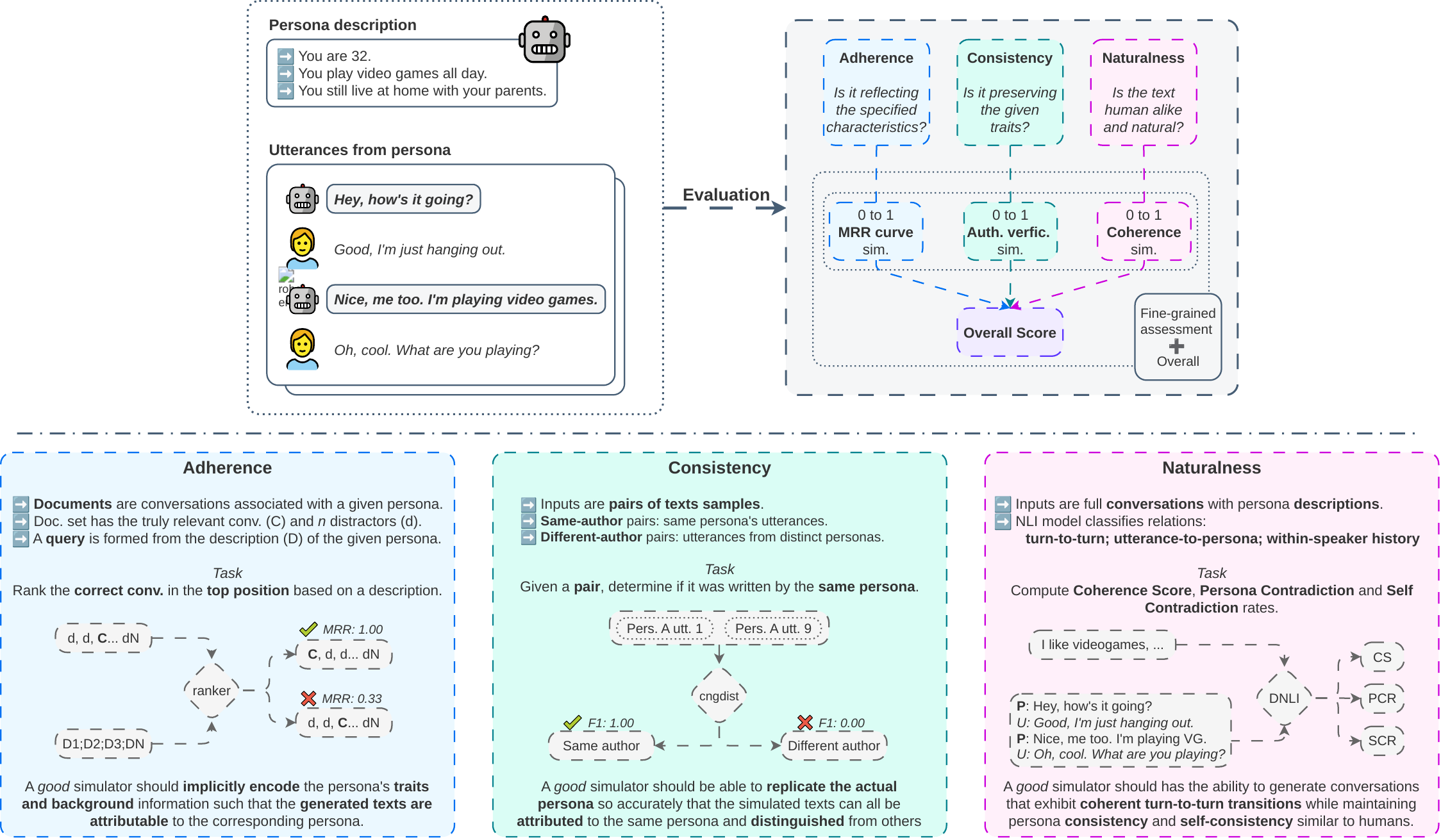}
    \caption{General overview of the \textsc{Eval4Sim} evaluation framework for persona simulation.}
    \Description{Diagram showing the \textsc{Eval4Sim} evaluation framework for persona simulation.}
    \label{fig:overview}
\end{figure*}

Persona grounding conditions a conversational agent on an explicit user profile~\cite{jung2016psychological,leary2011personality} specifying attributes, preferences, and behavioural tendencies. LLM personas are now widely used to simulate human conversations for user modelling, social reasoning, and behavioural analysis~\cite{cho2023integrative,wang2025user}, and ensuring that these simulations faithfully reflect human conversational behaviour has become a pressing evaluation challenge. A persona simulator must maintain a stable and distinguishable identity over time, expressing persona-specific cues consistently across dialogues~\cite{tseng2024two}, yet phenomena such as persona drift~\cite{li2024measuring}, inconsistency~\cite{li2025llm}, and exaggerated trait expression~\cite{bhandari2025can} limit their realism and validity. 

Existing evaluation approaches fall short in different ways. Traditional metrics such as BLEU~\cite{papineni2002bleu}, or perplexity~\cite{fang2025what} capture surface overlap rather than persona consistency~\cite{tseng2024two,liu2016how}. Specialized frameworks introduce persona-specific metrics~\cite{uehara2021fundamental}, consistency labels~\cite{welleck2019dialogue}, and human evaluation protocols~\cite{see2019what}, but isolate single attributes without considering trade-offs across dimensions. Recent benchmarks such as \textit{PersoBench}~\cite{afzoon2026persobench} and \textit{PersonaGym}~\cite{samuel2025personagym} advance the state of the art, yet both score individual turns against gold references or LLM-as-a-judge rubrics~\cite{zheng2023judging,desmond2025evalassist}, which are sensitive to prompt design and can exhibit systematic scoring biases~\cite{arabzadeh2025human-ai,dietz2025principles}. Crucially, none of these approaches grounds evaluation in empirical human behaviour: they assess quality relative to a gold response or a model opinion, not relative to how humans actually converse.

We address this gap with \textsc{Eval4Sim}\footnote{Code and documentation: \faicon{github}~\url{https://github.com/IRLab-UDC/eval4sim}.}, a corpus-agnostic framework for evaluating persona-grounded conversation simulation against a human reference corpus (Figure~\ref{fig:overview}). The framework jointly evaluates three dimensions: \emph{adherence}, whether persona traits are recoverable from dialogue; \emph{consistency}, whether speakers maintain a stable stylistic identity; and \emph{naturalness}, whether conversations follow human-like interaction patterns. Rather than rewarding maximal scores in isolation, \textsc{Eval4Sim} measures alignment to human behaviour and penalizes deviations in both directions. This is important because persona simulation involves trade-offs: making persona traits too recoverable may lead to unnatural self-disclosure, while optimizing fluent dialogue may weaken stylistic identity. \textsc{Eval4Sim} provides reproducible evaluations over ten simulation corpora, enabling researchers to diagnose where simulators diverge from human conversational patterns and to plug in new reference corpora or simulated datasets.
\section{The \textsc{Eval4Sim} Framework}
\label{sec:methodology}

\textsc{Eval4Sim} operates on two inputs: (i) a \emph{reference corpus} of human-to-human persona-grounded conversations that establishes the behavioural baseline, and (ii) one or more \emph{simulation corpora} to evaluate. Both contain multi-turn dialogues where each speaker is associated with a persona description. The framework is corpus-agnostic: any dataset with speaker-level persona annotations can serve as the reference. An expected input sample is shown below:

{
\tiny
\begin{CodeWithIcons}
(\textbf{Alex} \faicon{male}) User 1 persona traits:
    I like to dance at the club.
    I run a dog obedience school.
(\textbf{Emma} \faicon{female}) User 2 persona traits:
    I love to meet new people.
    I have a turtle named Timothy.

\textbf{Alex} (\faicon{male}) : Hi, I'm Alex. What's your name?
\textbf{Emma} (\faicon{female}) : Hi, I'm Emma. Nice to meet you.
\textbf{Alex} (\faicon{male}) : What are you interested in?
\textbf{Emma} (\faicon{female}): I enjoy meeting people and playing frisbee.
\end{CodeWithIcons}
}

Each dimension produces a human similarity score in $[0,1]$, where 1.0 indicates perfect alignment with the reference. The aggregate \textsc{e4s} score is the unweighted mean of the three, treating the dimensions as diagnostics of equal standing; practitioners with domain-specific priorities may reweight accordingly.

\paragraph{\textbf{1. Adherence via Dense Retrieval}}
\label{subsec:adherence}

We define \emph{adherence} as the ability of a simulator to implicitly expresses a persona's traits such that the generated conversation remains attributable to that persona. Formally, let $\mathcal{C}=\{c_1,\ldots,c_n\}$ be a set of conversations and $\mathcal{P}$ the set of personas, where each $p\in\mathcal{P}$ is associated with a set of relevant conversations $\mathcal{C}_p\subseteq\mathcal{C}$. Given $p$ as a query, the retrieval objective is to rank conversations in $\mathcal{C}_p$ above those in $\mathcal{C}\setminus\mathcal{C}_p$. We use ColBERT~\cite{khattab2020colbert} to retrieve conversations from persona descriptions, representing each conversation by the utterances of the evaluated speaker, since the query persona refers to that speaker. We evaluate MRR~\cite{craswell2009mean} over increasingly difficult pools, each containing one relevant conversation and $k$ distractors, yielding an MRR degradation curve per corpus.

Importantly, adherence should not be maximized blindly: a simulator may over-expose persona traits (e.g., ``As a teacher, I...''), making retrieval artificially easy but dialogue unnatural. We therefore compare each degradation curve to the human reference and penalizes deviations in either direction. Curves below the reference indicate under-encoding of persona cues, while curves above it indicate over-explicit persona signaling. Standard area-based scores are unsuitable because they reward systematically higher retrieval performance. Instead, we compute a scale-sensitive similarity between the simulation curve and the human reference curve:

{\small\begin{equation}
\text{Adh\textsubscript{sim}} = \frac{\sum_{i} w_i \cdot a_i \cdot b_i}{\max\left(\sum_{i} w_i \cdot a_i^2,\; \sum_{i} w_i \cdot b_i^2\right)}
\end{equation}}

Here, $a_i$ and $b_i$ are the MRR values at pool size $p_i$ for the reference and simulation, respectively, and $\delta_i$ are span weights accounting for unequal spacing between pool sizes. The score is $1$ when both curves match exactly and decreases when the simulated curve falls either below or above the human reference.

\paragraph{\textbf{2. Consistency via Authorship Verification}}
\label{subsec:consistency}

Whereas adherence measures whether persona \emph{content} is recoverable from the conversation, consistency measures whether each speaker maintains a stable \emph{style} distinguishable from other speakers. We define \emph{consistency} as the ability of a simulator to maintain a stable and distinguishable speaker identity across conversations: utterances from the same persona should exhibit coherent stylistic regularities while remaining separable from other personas. We implement this as authorship verification: given a pair of text samples $(t_1,t_2)$, the task is to predict whether both were produced by the same persona or by different personas. 

We use the character 4-gram TF--IDF baseline from PAN 2023~\cite{bevendorff2023overview}, which represents texts through TF--IDF weighted character $n$-grams, computes cosine similarities between pairs, and calibrates them into verification scores. This yields a \emph{separability} signal: if a simulator realizes personas consistently, same-persona pairs should be easier to distinguish from different-persona pairs. However, as with adherence, the goal is not to maximize separability. Excessive stylistic uniformity may make personas artificially easy to identify while producing repetitive, less human-like dialogue. We aggregate the five official PAN metrics (F1, AUC, Brier, c@1, F$_{0.5}^u$)~\footnote{We refer the reader to the official PAN overview~\cite{bevendorff2023overview} for full metric definitions.} into a single Consistency score. Human similarity penalizes deviation in either direction:

{\small\begin{equation}
\text{Con\textsubscript{sim}} = 1 - \frac{|\text{Consistency}_{\text{sim}} - \text{Consistency}_{\text{ref}}|}{\text{Consistency}_{\text{ref}}}
\end{equation}}

\paragraph{\textbf{3. Naturalness via NLI Distributions}}
\label{subsec:naturalness}

We define \emph{naturalness} as the ability of a simulator to generate conversations with human-like turn-to-turn dynamics while remaining compatible with the speaker's persona and internally self-consistent. We implement this dimension through dialogue-focused Natural Language Inference (NLI). Given a conversation with utterances $\{u_t\}_{t=1}^{T}$ and speaker-specific persona descriptions $\mathcal{P}_s=\{p_{s,1},\ldots,p_{s,n}\}$, we apply a DeBERTa-based NLI model fine-tuned on conversational data~\citep{welleck2019dialogue}\footnote{\texttt{zayn1111/deberta-v3-dnli}; 95.36\% accuracy on dialogue NLI.} to two types of pairs: (i) consecutive turn pairs $(u_{t-1},u_t)$, which capture local dialogue flow, and (ii) persona--utterance pairs $(p_{s,i},u_t)$, which capture compatibility between a speaker's utterances and their assigned persona. From these predictions, we derive three diagnostic metrics:

\begin{itemize}[leftmargin=*, noitemsep]
    \item \textbf{Coherence Score (CS):} the mean turn-to-turn coherence score over consecutive utterance pairs, assigning $1.0$ to entailment, $0.5$ to neutral, and $0.0$ to contradiction.
    \item \textbf{Persona Contradiction Rate (PCR):} the fraction of persona--utterance pairs predicted as contradiction with confidence at least $0.7$.\footnote{We use a confidence threshold to avoid counting uncertain NLI predictions as contradictions, since explicit persona contradictions are rare in well-formed human dialogues.}
    \item \textbf{Self-Contradiction Rate (SCR):} the fraction of within-speaker utterance pairs, computed within a 5-turn history window, predicted as contradiction with confidence at least $0.7$.
\end{itemize}

To obtain a compact Naturalness score, we aggregate CS, $(1-\mathrm{PCR})$, and $(1-\mathrm{SCR})$. We estimate the weights of each component from the human reference corpus: for each component, we compute its standard deviation across individual \textsc{PersonaChat} conversations and normalize the three standard deviations to sum to one. The intuition is that components with greater variability in natural dialogue provide more discriminative information, whereas near-constant components should contribute less. Since $\mathrm{std}(1-\mathrm{PCR})=\mathrm{std}(\mathrm{PCR})$ and $\mathrm{std}(1-\mathrm{SCR})=\mathrm{std}(\mathrm{SCR})$, the same weights apply to the non-contradiction terms. This yields weights of $0.565$ for CS, $0.126$ for $(1-\mathrm{PCR})$, and $0.309$ for $(1-\mathrm{SCR})$:

{\small\begin{equation}
\text{Naturalness} =
0.565 \times \mathrm{CS}
+ 0.126 \times (1-\mathrm{PCR})
+ 0.309 \times (1-\mathrm{SCR}).
\end{equation}}

Finally, as in the other dimensions, we convert this raw Naturalness score into a human-similarity score by comparing it with the reference corpus:

{\small\begin{equation}
\text{Nat\textsubscript{sim}} = 1 - \frac{|\text{Naturalness}_{\text{sim}} - \text{Naturalness}_{\text{ref}}|}{\text{Naturalness}_{\text{ref}}}
\end{equation}}
\section{Experiments and Results}
\label{sec:experiments}

\begin{table*}[!t]
    \centering
    \caption{Overall \textsc{e4s} evaluation. Similarity scores measure alignment to \textsc{PersonaChat} (1.000). \textsc{e4s} is the unweighted mean of the three similarity scores. \textbf{Bold}: 1st, \underline{underline}: 2nd, \textit{italic}: 3rd per column.}
    \label{tab:overall_results}
    \resizebox{0.75\textwidth}{!}{
    \begin{tabular}{lrrcrrcrrcrrcrr}
        \toprule
        & \multicolumn{3}{c}{\textbf{Adherence}} & \multicolumn{3}{c}{\textbf{Consistency}} & \multicolumn{3}{c}{\textbf{Naturalness}} & \multicolumn{3}{c}{\textbf{\textsc{e4s}}} \\
        \cmidrule(lr){2-4} \cmidrule(lr){5-7} \cmidrule(lr){8-10} \cmidrule(lr){11-13}
        \textbf{Dataset} & Score & Off. & Pos. & Score & Off. & Pos. & Score & Off. & Pos. & Score & Off. & Pos. \\
        \midrule
        \textit{PersonaChat} & \textit{1.000} & \textit{---} & \textit{---} & \textit{1.000} & \textit{---} & \textit{---} & \textit{1.000} & \textit{---} & \textit{---} & \textit{1.000} & \textit{---} & \textit{---} \\
        \midrule
        Qwen3 30B   & \underline{0.974} & -2.60\% & 2 & 0.917 & -8.30\% & 6 & \textbf{0.962} & -3.80\% & 1 & \textbf{0.951} & -4.90\% & 1 \\
        Gemma 3 12B & \textbf{0.983} & -1.70\% & 1 & 0.877 & -12.30\% & 8 & \textit{0.959} & -4.10\% & 3 & \underline{0.940} & -6.00\% & 2 \\
        Gemma 3 4B  & 0.948 & -5.20\% & 4 & 0.901 & -9.90\% & 7 & \underline{0.962} & -3.80\% & 2 & \textit{0.937} & -6.30\% & 3 \\
        Qwen3 4B    & 0.912 & -8.80\% & 7 & \textit{0.946} & -5.40\% & 4 & 0.951 & -4.90\% & 5 & 0.936 & -6.40\% & 4 \\
        Qwen3 14B   & 0.919 & -8.10\% & 5 & \textbf{0.976} & -2.40\% & 1 & 0.892 & -10.80\% & 9 & 0.929 & -7.10\% & 5 \\
        Gemma 3 27B & \textit{0.970} & -3.00\% & 3 & 0.874 & -12.60\% & 9 & 0.919 & -8.10\% & 7 & 0.921 & -7.90\% & 6 \\
        Gemma 3 1B  & 0.918 & -8.20\% & 6 & 0.867 & -13.30\% & 10 & 0.953 & -4.70\% & 4 & 0.913 & -8.70\% & 7 \\
        Qwen3 1.7B  & 0.840 & -16.00\% & 8 & 0.919 & -8.10\% & 5 & 0.932 & -6.80\% & 6 & 0.897 & -10.30\% & 8 \\
        SPC         & 0.811 & -18.90\% & 10 & \underline{0.955} & -4.50\% & 2 & 0.904 & -9.60\% & 8 & 0.890 & -11.00\% & 9 \\
        SPC-New     & 0.837 & -16.30\% & 9 & \textit{0.953} & -4.70\% & 3 & 0.874 & -12.60\% & 10 & 0.888 & -11.20\% & 10 \\
        \bottomrule
    \end{tabular}}
\end{table*}

\paragraph{\textbf{Experimental Setup}} We instantiate \textsc{Eval4Sim} using \textsc{Persona\\Chat}~\cite{zhang2018personalizing} as the reference corpus, a widely used collection of real persona-grounded human-to-human dialogues. We evaluate ten simulation corpora spanning two origins. The first two are published datasets: \textsc{Synthetic-Persona-Chat (SPC)}~\cite{jandaghi2024faithful}, which reuses \textsc{PersonaChat} persona descriptions with synthetic dialogues, and \textsc{SPC-New}~\cite{jandaghi2024faithful}, which uses fully synthetic personas and conversations. Both were produced with a Generator--Critic framework, where a generator creates candidate persona-grounded dialogues and a critic filters or selects them according to quality and persona-faithfulness criteria. The remaining eight corpora were generated in the experiments of this work by prompting open-source LLMs on the same \textsc{PersonaChat} personas: \textsc{Qwen3} (1.7B, 4B, 14B, 30B) and \textsc{Gemma~3} (1B, 4B, 12B, 27B). For each LLM, five independent conversation sets were generated under different random seeds. All reported scores are averaged at the corpus level across these five runs. Table~\ref{tab:datasets} summarises corpus statistics, while Table~\ref{tab:overall_results} reports the scores for all dimensions and the  \textsc{Eval4Sim} overall score.

\begin{table}[t]
      \centering
      \small
      \caption{Dataset statistics for \textsc{PersonaChat} and all simulations. Desc.: avg.\ items and words per description. Convs.: avg.\ messages and words per conversation.}
      \label{tab:datasets}
      \resizebox{0.88\columnwidth}{!}{
      \begin{tabular}{l c cc cc}
          \toprule
          \multirow{2}{*}{\textbf{Dataset}} &
          \multirow{2}{*}{\textbf{Samples}} &
          \multicolumn{2}{c}{\textbf{Desc.}} &
          \multicolumn{2}{c}{\textbf{Convs.}} \\
          \cmidrule(lr){3-4} \cmidrule(lr){5-6}
          & & $|\overline{items}|$ & $\overline{len.}$
            & $|\overline{mess.}|$ & $\overline{len.}$ \\
          \midrule
          \textsc{PersonaChat}~\cite{zhang2018personalizing} & 1,000  & 4.50 & 33.87 & 15.60 & 11.71 \\
          \textsc{SPC}~\cite{jandaghi2024faithful}              & 968    & 4.50 & 28.11 & 27.55 & 11.05 \\
          \textsc{SPC-New}~\cite{jandaghi2024faithful}       & 11,001 & 5.00 & 56.77 & 17.44 & 13.32 \\
          \hdashline
          \textsc{Qwen3 1.7B}                                & 645.4  & 4.50 & 28.13 & 29.83 & 12.29 \\
          \textsc{Qwen3 4B}                                  & 954.0  & 4.50 & 28.13 & 35.22 & 20.05 \\
          \textsc{Qwen3 14B}                                 & 946.6  & 4.50 & 28.11 & 31.27 & 22.34 \\
          \textsc{Qwen3 30B}                                 & 967.0  & 4.50 & 28.11 & 32.34 & 19.35 \\
          \textsc{Gemma 3 1B}                                & 743.2  & 4.50 & 28.02 & 25.47 & 20.71 \\
          \textsc{Gemma 3 4B}                                & 872.8  & 4.50 & 27.34 & 31.72 & 16.63 \\
          \textsc{Gemma 3 12B}                               & 964.0  & 4.50 & 28.11 & 32.37 & 18.94 \\
          \textsc{Gemma 3 27B}                               & 942.6  & 4.50 & 28.07 & 31.92 & 22.18 \\
          \bottomrule
      \end{tabular}}
\end{table}

\paragraph{\textbf{Adherence}}

\begin{figure}[t]
    \centering
    \includegraphics[width=0.88\linewidth]{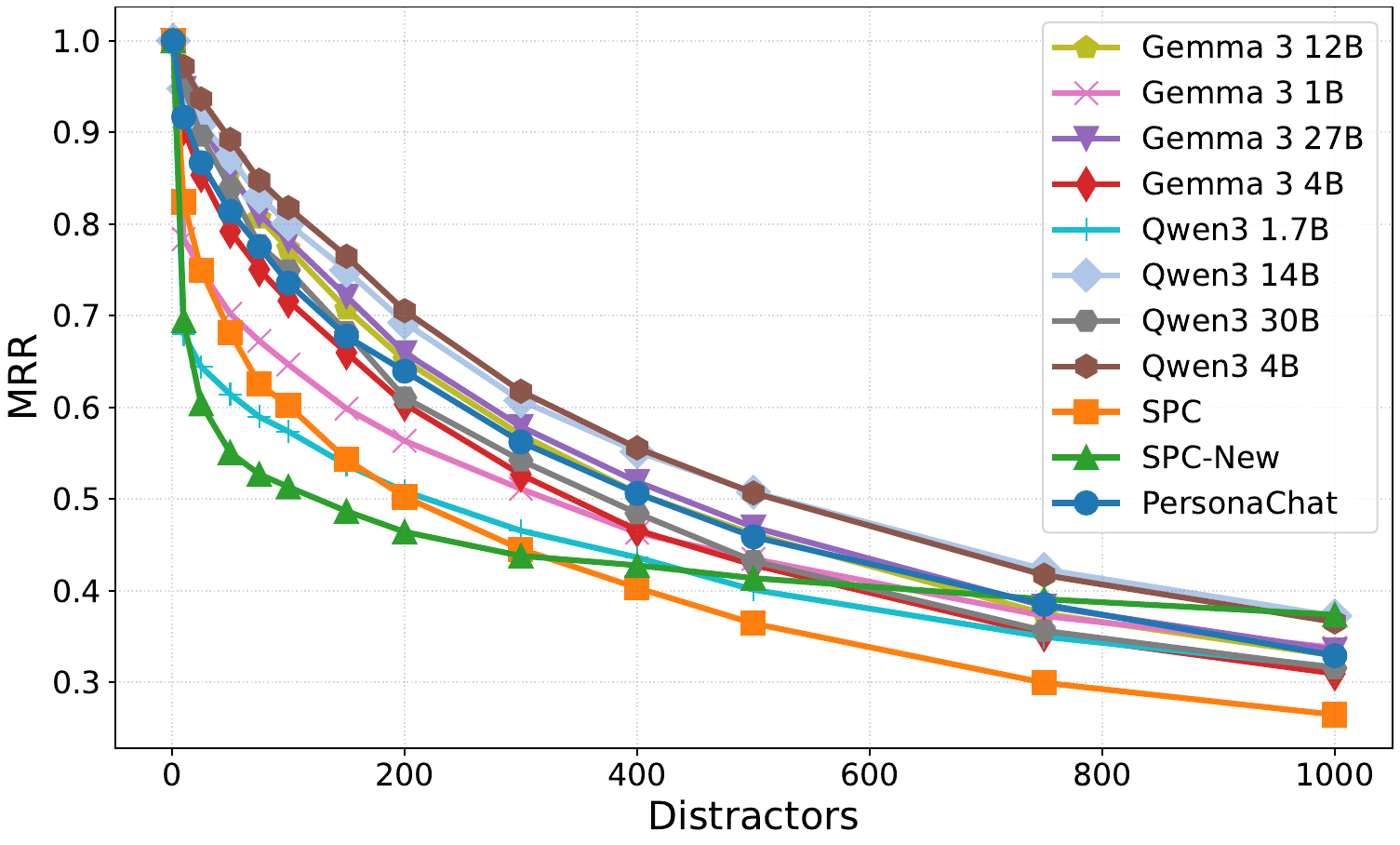}
    \caption{MRR degradation across candidate pool sizes for \textsc{PersonaChat} and ten simulated datasets.}
    \Description{Line plot showing MRR on the y-axis versus pool size on the x-axis for eleven datasets.}
    \label{fig:adherence_plot}
\end{figure}

Figure~\ref{fig:adherence_plot} shows MRR degradation as the distractor pool grows. \textsc{PersonaChat} drops from near-perfect retrieval at small pools to roughly 0.3 at 1000 distractors. Several LLM-generated corpora closely track this human pattern, suggesting that their persona cues are recoverable without becoming unnaturally explicit. The Generator--Critic datasets deviate more clearly: \textsc{SPC} remains consistently below the human curve, indicating weaker persona attribution, while \textsc{SPC-New} shows a sharp early drop before partially converging. Overall, direct generation with modern LLMs yields more human-aligned persona encoding than the older Generator--Critic corpora. \textsc{Gemma~3~12B} is closest to the human baseline ($-1.70\%$), followed by \textsc{Qwen3~30B} ($-2.60\%$). \textsc{SPC} and \textsc{SPC-New} are least aligned ($-18.90\%$ and $-16.30\%$).

\paragraph{\textbf{Consistency}}

The datasets closest to \textsc{PersonaChat} under the stylometry-based verifier are \textsc{Qwen3~14B} ($-2.40\%$), \textsc{SPC} ($-4.50\%$), and \textsc{SPC-New} ($-4.70\%$). This contrasts with adherence, where larger modern LLMs perform best, showing that model capacity alone does not guarantee human-like stylistic stability. Notably, \textsc{Qwen3~1.7B} attains a higher raw consistency score than \textsc{PersonaChat} but lower similarity (0.919), illustrating the bidirectional nature of the metric: the goal is not maximal verifier performance, but matching the human level of stylistic separability.

\paragraph{\textbf{Naturalness}}

\textsc{Qwen3~30B} and \textsc{Gemma~3~4B} lead in naturalness (both $-3.80\%$), with the top five systems within 1.1\% of the human baseline. The clearest trade-off is \textsc{Qwen3~14B}: it ranks first in consistency but ninth in naturalness ($-10.80\%$), suggesting that highly distinguishable stylistic identity may come at the cost of human-like dialogue flow. The Generator--Critic datasets rank near the bottom: \textsc{SPC-New} is least aligned ($-12.60\%$), consistent with overly regular conversational flow relative to the human baseline. Since the Naturalness weights are estimated from variability in \textsc{PersonaChat}, these results should be interpreted as alignment to the baseline rather than absolute conversational quality.

\paragraph{\textbf{Overall: \textsc{e4s}}}

Table~\ref{tab:overall_results} confirms that no simulator matches human behaviour uniformly across dimensions. \textsc{Qwen3~30B} ranks first overall (0.951), combining strong adherence (rank~2) with the best naturalness alignment (rank~1), at the cost of weaker consistency (rank~6). The Generator--Critic datasets show the opposite profile: competitive consistency but the weakest adherence and poor naturalness, resulting in the lowest overall scores. \textsc{Qwen3~14B} achieves the best consistency yet ranks ninth on naturalness, while \textsc{Gemma~3~4B} outperforms \textsc{Gemma~3~27B}, confirming that scale alone does not determine alignment. These trade-offs reinforce the need for multi-dimensional evaluation: optimizing one dimension can move a simulator away from the human baseline on others.
\section{Resource Description and Conclusions}
\label{sec:conclusion}

\textsc{Eval4Sim} is publicly released at \url{https://github.com/IRLab-UDC/eval4sim}. As a corpus-agnostic resource, users can replace \textsc{PersonaChat} with any persona-annotated conversational corpus and evaluate new simulators against the corresponding human reference. Experiments over ten simulation corpora illustrate its diagnostic value. First, LLM-generated corpora achieve stronger adherence alignment than prior Generator--Critic datasets, while consistency follows a different ordering: larger models do not necessarily yield better stylistic alignment. Second, naturalness results show that several simulators deviate from the human reference through overly regular conversational flow, meaning high apparent coherence does not always imply human-like dialogue. Third, no simulator matches human behaviour uniformly across dimensions: \textsc{Qwen3~30B} achieves the best overall balance, but still trades off consistency against adherence and naturalness. These findings underscore the need for multi-dimensional evaluation, since optimizing a single aspect of persona simulation can move a system away from the human baseline on others.

\begin{acks}
The first author acknowledges the support of the Department of Education, Science, Universities, and Vocational Training of the Xunta de Galicia (grant ED481A-2024-079). All authors affiliated with IRLab and CITIC acknowledge funding from the Ministry of Science, Innovation and Universities of the Government of Spain (projects PID2022-137061OB-C21, PID2025-167749OB-C22), as well as from the Department of Education, Science, Universities, and Vocational Training of the Xunta de Galicia (grant GRC ED431C 2025/49). CITIC, as a center accredited for excellence within the Galician University System and a member of the CIGUS Network, receives subsidies from the Department of Education, Science, Universities, and Vocational Training of the Xunta de Galicia. Additionally, it is co-financed by the EU through the FEDER Galicia 2021-27 operational program (Ref. ED431G 2023/01).
\end{acks}

\section*{GenAI Usage Disclosure}
During the preparation of this manuscript, generative AI tools were employed solely for light editing purposes, including proofreading, grammar correction, vocabulary improvement, and overall language polishing. All substantive ideas, analyses, experiments, and written content were created by the co-authors without direct text generation from any AI model.

\balance

\bibliographystyle{ACM-Reference-Format}
\bibliography{references}

\end{document}